\pdfoutput=1
\documentclass[letterpaper]{article} 
\usepackage[preprint,nonatbib]{nips_2018}  
\usepackage{booktabs}
\usepackage{times}  
\usepackage{helvet}  
\usepackage{courier}  
\usepackage{url}  
\usepackage{graphicx}  
\usepackage{rotating}
\usepackage{wrapfig}
\usepackage{multicol}
\usepackage{latexsym}
\usepackage{amsmath}
\usepackage{amssymb}
\usepackage{url}
\usepackage[noend]{algorithmic}
\usepackage{algorithm}
\usepackage{multirow}
\usepackage{mathtools}
\usepackage{enumitem}
\usepackage{color}
\usepackage{microtype}
\usepackage{tabularx}
\usepackage{relsize}
\usepackage{amsthm}
\usepackage{bm}
\usepackage{amsthm}
\usepackage{epstopdf}

\newtheoremstyle{mytheoremstyle} 
{\topsep}                    
{\topsep}                    
{\itshape}                   
{}                           
{\scshape}                   
{.}                          
{.5em}                       
{\thmname{#1}\thmnumber{ #2}\thmnote{ (#3)}}  

\theoremstyle{mytheoremstyle}

\newtheorem{thm}{Theorem}

\newtheorem{lem}{Lemma}

\newtheorem{defn}{Definition}

\newcommand{\Real}{\mathbb R}

\newcommand{\R}{\mathcal{R}}
\newcommand{\X}{\mathbb{X}}
\newcommand{\Xcal}{\mathcal{X}}

\newcommand{\Ycal}{\mathcal{Y}}
\newcommand{\Z}{\mathbb{Z}}

\newcommand{\iirl}{\textsf{I2RL}}

\newcommand{\irl}{\textsc{IRL}}

\begin{document}

\title{A Framework and Method for Online Inverse Reinforcement Learning}  

\author{Saurabh Arora\\
  Department of Computer Science\\
  University of Georgia, Athens GA 30605\\
  \texttt{sa08751@uga.edu}
  \And
  Prashant Doshi\\
  Department of Computer Science\\
  University of Georgia, Athens GA 30605\\
  \texttt{pdoshi@cs.uga.edu}
  \And
  Bikramjit Banerjee\\
  School of Computing\\
  University of Southern Mississippi, Hattiesburg, MS 39406\\
  \texttt{Bikramjit.Banerjee@usm.edu}
}  

\maketitle

\begin{abstract}  
  Inverse reinforcement learning (IRL) is  the problem of learning the
  preferences of an  agent from the observations of its  behavior on a
  task. While  this problem  has been  well investigated,  the related
  problem   of  {\em   online}   IRL---where   the  observations   are
  incrementally  accrued, yet  the  demands of  the application  often
  prohibit a  full rerun  of an  IRL method---has  received relatively
  less attention. We  introduce the first formal  framework for online
  IRL, called {\em  incremental IRL} (\iirl{}), and a  new method that
  advances maximum entropy IRL with hidden variables, to this setting.
  Our formal  analysis shows that  the new method has  a monotonically
  improving  performance  with more  demonstration  data,  as well  as
  probabilistically  bounded  error,  both   under  full  and  partial
  observability.  Experiments  in a  simulated robotic  application of
  penetrating a continuous patrol under occlusion shows the relatively
  improved performance  and speed up  of the new method  and validates
  the utility of online IRL.
\end{abstract}


\vspace{-0.05in}
\section{Introduction}
\label{sec:intro}
\vspace{-0.05in}

Inverse                     reinforcement                     learning
(IRL)~\cite{Ng00:Algorithms,Russell98:Learning} refers  to the problem
of  ascertaining  an  agent's  preferences from  observations  of  its
behavior on  a task.   It inverts  RL with its  focus on  learning the
reward function  given information about optimal  action trajectories.
IRL lends itself naturally to  a robot learning from demonstrations by
a human teacher (often called  the expert) in controlled environments,
and   therefore    finds   application   in   robot    learning   from
demonstration~\cite{Argall09:Survey}           and          imitation
learning~\cite{Osa18:Algorithmic}.

Previous                          methods                          for
IRL~\cite{Abbeel04:Apprenticeship,Babes-Vroman11:Apprenticeship,Boularias12:Structured,Choi11:Inverse,Ramachandran07:Bayesian,Ziebart08:Maximum}
typically  operate  on large  batches  of  observations and  yield  an
estimate of the  expert's reward function in a  one-shot manner. These
methods fill  the need  of applications  that predominantly  center on
imitation learning.   Here, the task  being performed is  observed and
must be  replicated subsequently.  However, newer  applications of IRL
are motivating the need for continuous learning from streaming data or
data in mini-batches.  Consider, for  example, the task of forecasting
a person's goals in an  everyday setting from observing his activities
using a  body camera~\cite{Rhinehart17:First}. Alternately,  a robotic
learner observing  continuous patrols from  a vantage point  is tasked
with penetrating the patrolling and  reaching a goal location speedily
and   without   being   spotted~\cite{Bogert14:Multi}.    Both   these
applications  offer streaming  observations,  no  episodic tasks,  and
would  benefit from  progressively  learning and  assessing the  other
agent's preferences.

In this  paper, we  present the first  formal framework  to facilitate
investigations into online IRL.  The framework, labeled as incremental
IRL  (\iirl{}), establishes  the key  components of  this problem  and
rigorously defines the notion of an incremental variant of IRL.  Next,
we   focus   on   methods   for  online   IRL.    Very   few   methods
exist~\cite{Jin10:Convergence,Rhinehart17:First}  that are  suited for
online  IRL, and  we  cast these  in the  formal  context provided  by
\iirl{}. More importantly, we introduce  a new method that generalizes
recent advances in maximum entropy IRL with hidden training data to an
online setting. Key theoretical properties of this new method are also
established. In particular, we show that the performance of the method
improves   monotonically   with   more    data   and   that   we   may
probabilistically bound the error in  estimating the true reward after
some data both under full observability and when some data is hidden.

Our experiments evaluate  the benefit of online IRL  on the previously
introduced robotic application of penetrating continuous patrols under
occlusion~\cite{Bogert14:Multi}.  
We comprehensively  demonstrate that  the proposed  incremental method
achieves  a  learning performance  that  is  similar  to that  of  the
previously introduced batch  method.  More importantly, it  does so in
significantly less time thereby suffering  from far fewer timeouts and
a significantly improved success  rate. Consequently, this paper makes
important initial  contributions toward the nascent  problem of online
IRL by  offering both a formal  framework, \iirl{}, and a  new general
method that performs well.



\vspace{-0.05in}
\section{Background on IRL}
\label{sec:background}
\vspace{-0.05in}

Informally, IRL  refers to  both the  problem and  method by  which an
agent learns  preferences of another  agent that explain  the latter's
observed  behavior~\cite{Russell98:Learning}.   Usually considered  an
``expert'' in the task that it  is performing, the observed agent, say
$E$, is  modeled as  executing the  optimal policy  of a  standard MDP
defined as $\langle S_E,A_E,T_E,R_E  \rangle$.  The learning agent $L$
is assumed  to perfectly  know the  parameters of  the MDP  except the
reward function.   Consequently, the learner's  task may be  viewed as
finding a reward  function under which the  expert's observed behavior
is optimal.

This problem  in general is  ill-posed because for any  given behavior
there  are  infinitely-many  reward  functions which  align  with  the
behavior.  Ng and Russell~\cite{Ng00:Algorithms} first formalized this
task as a  linear program in which the reward  function that maximizes
the difference in value between the  expert's policy and the next best
policy  is   sought.   Abbeel   and  Ng~\cite{Abbeel04:Apprenticeship}
present  an algorithm  that  allows  the expert  $E$  to provide  task
demonstrations instead of its policy.   The reward function is modeled
as   a  linear   combination   of  $K$   binary  features,   $\phi_k$:
$S_E \times A_E$  $\rightarrow$ $[0,1]$ $ , k \in  \{1,2 \ldots K\} $,
each of which maps a state from  the set of states $S_E$ and an action
from the set  of $E$'s actions $A_E$  to a value in  [0,1].  Note that
non-binary  feature  functions can  always  be  converted into  binary
feature functions although there will be more of them. Throughout this
article, we assume that these features are known to or selected by the
learner.   The reward  function  for  expert $E$  is  then defined  as
$R_E(s,a)  =  \bm{\theta}^T  \phi(s,a)  =  \sum_{k=1}^K  \theta_k\cdot
\phi_k(s,a)$,  where  $\theta_k$  are  the  {\em  weights}  in  vector
$ \bm{\theta} $; let $\R = \Real^{|S_E \times A_E|}$ be the continuous
space of the reward
functions. 
The learner's  task is  reduced to  finding a  vector of  weights that
complete the reward  function, and subsequently the MDP  such that the
demonstrated behavior is optimal.

To assist in finding the  weights, feature expectations are calculated
for  the expert's  demonstration  and compared  to  those of  possible
trajectories~\cite{Ziebart08:Maximum}.  A demonstration is provided as
one or  more {\em  trajectories}, which are  a sequence  of length-$T$
state-action                                                    pairs,
$( \langle s, a \rangle_1, \langle  s, a \rangle_2, \dots \langle s, a
\rangle_T )$, corresponding to an observation of the expert's behavior
across  $T$  time  steps.   Feature expectations  of  the  expert  are
discounted     averages     over    all     observed     trajectories,
$\hat{\phi}_k        =         \frac{1}{|X|}        \sum_{X        \in
  \Xcal{}} \sum\nolimits_{t=1}^T \gamma^t~\phi_k(\langle  s,  a  \rangle_t)$,
where $X$  is a trajectory  in the set of all  observed trajectories,
$\Xcal{}$, and $\gamma \in (0,1)$ is a discount factor.
%
%
Given a set of reward weights the expert's MDP is completed and solved
optimally      to     produce      $\pi^*_E$.      The      difference
$\hat{\phi} - \phi^{\pi^*_E}$ provides a  gradient with respect to the
reward weights for  a numerical solver.  


\vspace{-0.05in}
\subsection{Maximum Entropy IRL}
\label{subsec:maxent}
\vspace{-0.05in}

While expected to  be valid in some contexts,  the max-margin approach
of Abeel  and Ng~\cite{Abbeel04:Apprenticeship} introduces a  bias into
the learned reward  function in general.  To address  this, Ziebart et
al.~\cite{Ziebart08:Maximum} find  the distribution with  {\em maximum
  entropy}  over all  trajectories that  is constrained  to match  the
observed feature expectations.

\begin{align}
\begin{array}{l}
  \max \limits_{\Delta} \left( -\sum\nolimits_{X \in \mathbb{X}} Pr(X)~ log~Pr(X) \right )\\\\
  \mbox{{\bf subject to}}~~~
  \sum \nolimits_{X \in \mathbb{X}} Pr(X) = 1 \\
  \sum \nolimits_{X \in \mathbb{X}} Pr(X)~\sum_{t=1}^T \gamma^t \phi_k(\langle s, a \rangle_t)  = \hat{\phi}_k ~~~~~~ \forall k 
\label{eq:ziebart-max-ent}
\end{array}
\end{align}

Here,  $\Delta$ is  the  space  of all  distributions  over the  space
$ \mathbb{X} $ of all  trajectories.  We denote the analytical feature
expectation on  the left hand side  of the second constraint  above as
$E_{\X}[\phi_k]$.
Equivalently, as the distribution  is parameterized by learned weights
$\bm{\theta}$, $E_{\X}[\phi_k]$ represents the feature expectations of
policy  $\pi^*_E$ computed  using  the learned  reward function.   The
benefit is  that distribution $  Pr(X;\bm{\theta}) $ makes  no further
assumptions beyond those which are needed to match its constraints and
is maximally noncommittal to any one  trajectory.  As such, it is most
generalizable by being  the least wrong most often  of all alternative
distributions.   {\em  A disadvantage  of  this  approach is  that  it
  becomes  intractable  for  long  trajectories  because  the  set  of
  trajectories  grows exponentially  with  length.}   In this  regard,
another  formulation defines  the  maximum  entropy distribution  over
policies~\cite{Boularias12:Structured},  the  size  of which  is  also
large but fixed.

\vspace{-0.05in}
\subsection{IRL under Occlusion}
\label{subsec:occlusion}
\vspace{-0.05in}

Our motivating application involves a  subject robot that must observe
other mobile robots from a fixed  vantage point.  Its sensors allow it
a limited observation area; within this  area it can observe the other
robots fully,  outside this area  it cannot observe at  all.  Previous
methods~\cite{Bogert14:Multi,Bogert15:Toward} denote this special case
of  partial  observability  where  certain  states  are  either  fully
observable or  fully hidden as \textit{occlusion}.   Subsequently, the
trajectories gathered  by the learner exhibit  missing data associated
with  time steps  where the  expert robot  is in  one of  the occluded
states.    The   empirical   feature   expectation   of   the   expert
$\hat{\phi}_k$ will therefore exclude the occluded states (and actions
in those states).

To ensure that the feature  expectation constraint in IRL accounts for
the  missing data,  Bogert  and Doshi~\cite{Bogert14:Multi}  while
maximizing  entropy over  policies~\cite{Boularias12:Structured} limit
the  calculation of  feature expectations  for policies  to observable
states
only.  
%
A recent approach~\cite{Bogert16:Expectation}  improves on this method
by  taking an  expectation over  the missing  data conditioned  on the
observations.  Completing the missing data  in this way allows the use
of all  states in  the constraint  and with  it the  Lagrangian dual's
gradient as well.  The nonlinear program in~\eqref{eq:ziebart-max-ent}
is modified to account for the hidden data and its expectation.

Let $Y$  be the observed  portion of a  trajectory, $Z$ is one  way of
completing the hidden portions of  this trajectory, $\Z$ is the set of
all possible  $Z$, and $X =  Y \cup Z$.  Now  we may treat  $Z$ as a
latent variable and take the expectation to arrive at a new definition
for the expert's feature expectations:
\begin{align}
  \hat{\phi}^{Z|_Y}_k  \triangleq  \frac{1}{|\Ycal{}|}  \sum\limits_{Y
    \in   \Ycal}   \sum\limits_{Z   \in   \Z}   Pr(Z|Y;\bm{\theta})
\sum\nolimits_{t=1}^T \gamma^t \phi_k(\langle s, a \rangle_t)
\label{eq:latent-phi}
\end{align}
where $\langle  s,a \rangle_t \in Y  \cup Z$, $\Ycal{}$ is  the set of
all  observed   $Y$  and  $\Xcal{}$   is  the  set  of   all  complete
trajectories. The program in~\eqref{eq:ziebart-max-ent} is modified by
replacing  $\hat{\phi}_k$  with   $\hat{\phi}^{Z|_Y}_k$,  as  we  show
below.  Notice that  in the  case of  no occlusion  $\Z$ is  empty and
$\Xcal{} =  \Ycal{}$.  Therefore $\hat{\phi}^{Z|_Y}_k  = \hat{\phi}_k$
and  this method  reduces  to~\eqref{eq:ziebart-max-ent}.  Thus,  this
method generalizes the previous maximum entropy IRL method.
\begin{align}
\begin{array}{l}
 \max \limits_{\Delta} \left( -\sum\nolimits_{X \in \X} Pr(X)~ log~Pr(X) \right )\\\\
 \mbox{{\bf subject to}}~~~
 \sum \nolimits_{X \in \X} Pr(X) = 1 \\
 \sum \nolimits_{X \in \X} Pr(X)~
 \sum_{t=1}^T \gamma^t \phi_k(\langle s, a \rangle_t)
  = \hat{\phi}^{Z|_Y}_k ~~~~~~ \forall k 
\end{array}
\label{eq:EM-max-ent}
\end{align}

However, the program in~\eqref{eq:EM-max-ent} becomes nonconvex due to
the presence of $Pr(Z|Y)$.  As  such, finding its optima by Lagrangian
relaxation is not trivial.   Wang et al.~\cite{Wang12:Latent} suggests
a log linear approximation to  obtain maximizing $Pr(X)$ and casts the
problem of finding the reward  weights as likelihood maximization that
can       be       solved       within       the       schema       of
expectation-maximization~\cite{Dempster77:EM}.  An application of this
approach to  the problem of  IRL under occlusion yields  the following
two steps with more details in~\cite{Bogert16:Expectation}:

\noindent    {\bf    E-step}    This   step    involves    calculating
Eq.~\ref{eq:latent-phi}  to  arrive  at  $\hat{\phi}^{Z|_Y,(t)}_k$,  a
conditional  expectation  of  the  $K$  feature  functions  using  the
parameter  $\bm{\theta}^{(t)}$ from  the  previous  iteration. We  may
initialize the parameter vector
randomly. 

\noindent    {\bf    M-step}    In    this    step,    the    modified
program~\eqref{eq:EM-max-ent}     is     optimized    by     utilizing
$\hat{\phi}^{Z|_Y,(t)}_k$  from  the  E-step  above  as  the  expert's
constant   feature  expectations   to  obtain   $\bm{\theta}^{(t+1)}$.
Normalized                    exponentiated                   gradient
descent~\cite{Steinhardt14:Adaptivity} solves the program.

As  EM may converge  to local  minima, this  process is  repeated with
random initial $\bm{\theta}$ and the solution with the maximum entropy
is chosen as the final one.

\vspace{-0.05in}
\section{Incremental IRL (\iirl)}
\label{sec:i2rl}
\vspace{-0.05in}

We present our framework labeled \iirl{} in order to realize IRL in an
online setting.  Then, we present an  initial set of methods cast into
the framework of \iirl{}. In addition to including previous techniques
for online IRL, we introduce a new method that generalizes the maximum
entropy IRL involving latent variables. 

\vspace{-0.05in}
\subsection{Framework}
\label{subsec:defn}
\vspace{-0.05in}

Expanding  on  the  notation   and  definitions  given  previously  in
Section~\ref{sec:background},  we introduce  some new  notation, which
will help us in defining \iirl{} rigorously.

\begin{defn}[Set  of   fixed-length  trajectories]  The  set   of  all
  trajectories  of finite  length $T$  from an  MDP attributed  to the
  expert            $E$            is           defined            as,
  $\X^T  = \{X|X=(  \langle s,  a \rangle_1,  \langle s,  a \rangle_2,
  \ldots \langle  s, a \rangle_T ),  \forall s \in S_E,  \forall a \in
  A_E \}$.
\end{defn}

Let $\mathbb{N}^+$ be a bounded set  of natural numbers. Then, the set
of                 all                 trajectories                 is
$ \mathbb{X} =  {\X^1 \cup \X^2 \cup  \ldots \cup \X^{|\mathbb{N}^+|}}
$.  Recall that a demonstration is  some finite set of trajectories of
varying lengths, $\Xcal{} = \{X^T|X^T \in \X^T, T \in \mathbb{N}^+\}$,
and it  includes the  empty set.~\footnote{Repeated trajectories  in a
  demonstration  can  usually be  excluded  for  many methods  without
  impacting  the learning.}  Subsequently, we  may define  the set  of
demonstrations.

 \begin{defn}[Set of demonstrations] The  set of demonstrations is the
   set  of  all subsets  of  the  spaces  of trajectories  of  varying
   lengths.      Therefore,      it     is     the      power     set,
   $ 2^\mathbb{X}    =   2^{\X^1    \cup    \X^2    \cup   \ldots    \cup
     \X^{|\mathbb{N}^+|}}$.
\end{defn}

In the context of the definitions above, traditional IRL attributes an
MDP without  the reward function  to the expert, and  usually involves
determining   an   estimate   of   the   expert's   reward   function,
$\hat{R}_E \in  \R$, which  best explains the  observed demonstration,
$\Xcal{}  \in  2^\X$.   As  such,  we may  view  IRL  as  a  function:
$\zeta (MDP_{/R_E},\Xcal{}) = \hat{R}_E$.

To establish  the definition of \iirl{},  we must first define  a {\em
  session} of \iirl{}. Let $\hat{R}^0_E$ be an initial estimate of the
expert's reward function.

\begin{defn}[Session] A  session of  \iirl{} represents  the following
  function:
  $\zeta_i(MDP_{/R_E},\Xcal{}_i,\hat{R}^{i-1}_E)  = \hat{R}^i_E$.   In
  this $i^{th}$  session where  $i >  0$, \iirl{}  takes as  input the
  expert's MDP  sans the  reward function, the  demonstration observed
  since the  previous session,  $\Xcal{}_i \in  2^\X$, and  the reward
  function estimate  learned from  the previous  session. It  yields a
  revised estimate of the expert's rewards from this session of IRL.
\label{def:session}
\end{defn}
Note  that we  may replace  the  reward function  estimates with  some
statistic that is sufficient to compute it (e.g., $ \bm{\theta} $).

We may let the sessions  run infinitely. Alternately, we may establish
stopping criteria for  \iirl{}, which would allow  us to automatically
terminate  the   sessions  once  the  criterion   is  satisfied.   Let
$LL(\hat{R}_E^i|\Xcal{}_{1:i})$   be  the   log   likelihood  of   the
demonstrations received up  to the $i^{th}$ session  given the current
estimate of the expert's reward function.  We may view this likelihood
as a  measure of  how well  the learned  reward function  explains the
observed data.

\begin{defn}[Stopping criterion \#1] Terminate  the sessions of \iirl{}
  when
  $|LL(\hat{R}_E^i|\Xcal{}_{1:i})                                    -
  LL(\hat{R}_E^{i-1}|\Xcal{}_{1:i-1})|   \leqslant  \epsilon$,   where
  $\epsilon$ is a very small positive number and is given.
\label{def:stop1}
\end{defn}
Definition~\ref{def:stop1} reflects the  fact that additional sessions
are not  impacting the  learning significantly. On  the other  hand, a
more effective stopping criterion is  possible if we know the expert's
true    policy.     We    utilize   the    {\em    inverse    learning
  error}~\cite{Choi11:Inverse} in this criterion, which gives the loss
of value  if $L$ uses  the learned policy on  the task instead  of the
expert's:  $ILE(\pi^*_E,\pi_E^L)  =  ||V^{\pi^*_E}-  V^{\pi_E^L}||_1$.
Here, $V^{\pi^*_E}$  is the  optimal value function  of $E$'s  MDP and
$V^{\pi_E^L}$  is the  value  function due  to  utilizing the  learned
policy $\pi_E^L$  in $E$'s MDP.   Notice that when the  learned reward
function  results in  an  identical optimal  policy  to $E$'s  optimal
policy,  $\pi^*_E  =   \pi_E^L$,  ILE  will  be   zero;  it  increases
monotonically as the two policies increasingly diverge in value.

\begin{defn}[Stopping criterion \#2] Terminate the sessions of \iirl{}
  when
  $ILE(\pi^*_E,\pi_E^{i-1})    -     ILE(\pi^*_E,\pi_E^i)    \leqslant
  \epsilon$, where  $\epsilon$ is a  very small positive error  and is
  given. Here, $\pi_E^i$  is the optimal policy  obtained from solving
  the expert's MDP  with the reward function  $\hat{R}_E^i$ learned in
  session $i$.
\label{def:stop2}
\end{defn}
Obviously,   prior   knowledge  of   the   expert's   policy  is   not
common. Therefore, we view this  criterion as being more useful during
the formative assessments of \iirl{} methods.

Utilizing                    Defs.~\ref{def:session},~\ref{def:stop1},
and~\ref{def:stop2}, we formally define \iirl{} next.
\begin{defn}[\iirl{}]  Incremental  IRL  (\iirl{}) is  a  sequence  of
  learning                                                    sessions
  $\{\zeta_1(MDP_{/R_E},\Xcal{}_1,\hat{R}^0_E),
  \zeta_2(MDP_{/R_E},\Xcal{}_2,\hat{R}^1_E),
  \zeta_3(MDP_{/R_E},\Xcal{}_3,\hat{R}^2_E),\ldots,\}$, which continue
  infinitely, or  until either  stopping criterion \#1  or \#2  is met
  depending on which one is chosen a'priori.
\end{defn}

\vspace{-0.05in}
\subsection{Methods}
\label{subsec:methods}
\vspace{-0.05in}

Our goal  is to facilitate  a portfolio of  methods each with  its own
appealing properties under the framework of \iirl{}. This would enable
online  IRL  in  various  applications. An  early  method  for  online
IRL~\cite{Jin10:Convergence}   modifies   Ng  and   Russell's   linear
program~\cite{Ng00:Algorithms} to  take as input a  single trajectory
(instead  of  a  policy)  and  replaces the  linear  program  with  an
incremental update of the reward  function. We may easily present this
method  within the  framework of  \iirl{}.  A  session of  this method
$\zeta_i(MDP_{/R_E},\Xcal{}_i,\hat{R}_E^{i-1})$    is   realized    as
follows:   Each   $\Xcal{}_i$   is    a   single   state-action   pair
$\langle      s,a      \rangle$;     initial      reward      function
$\hat{R}_E^0   =  \frac{1}{\sqrt{|S_E|}}$,   whereas  for   $i  >   0$
$\hat{R}_E^i = \hat{R}_E^{i-1} + \alpha\cdot  v_i$, where $v_i$ is the
difference in expected  value of the observed action $a$  at state $s$
and the  (predicted) optimal action  obtained by solving the  MDP with
the reward  function $\hat{R}_E^{i-1}$,  and $\alpha$ is  the learning
rate.   While  no  explicit   stopping  criterion  is  specified,  the
incremental   method  terminates   when  it   runs  out   of  observed
state-action  pairs. Jin  et al.~\cite{Jin10:Convergence}  provide the
algorithm for this method as well as error bounds.

A  recent  method  by  Rhinehart  and  Kitani~\cite{Rhinehart17:First}
performs online IRL  for activity forecasting. Casting  this method to
the   framework   of   \iirl{},   a  session   of   this   method   is
$\zeta_i(MDP_{/R_E},\Xcal{}_i,\bm{\theta}^{i-1})$,     which    yields
$\bm{\theta}^i$. Input observations for  the session, $\Xcal{}_i$, are
all the activity trajectories observed  since the end of previous goal
until  the next  goal is  reached. The  session IRL  finds the  reward
weights     $\bm{\theta}^i$      that     minimize      the     margin
$\phi^{\pi^*_E}  -  \hat{\phi}$  using gradient  descent.   Here,  the
expert's policy  $\pi^*_E$ is obtained  by using soft  value iteration
for solving the complete MDP  that includes a reward function estimate
obtained  using  previous  weights $\bm{\theta}^{i-1}$.   No  explicit
stopping  criterion  is  utilized  for  the  online  learning  thereby
emphasizing its continuity.

\subsubsection{Incremental Latent MaxEnt} 

We present a new method for online IRL under the I2RL framework, which
modifies   the   latent   maximum  entropy   (LME)   optimization   of
Section~\ref{subsec:occlusion}.  It  offers the capability  to perform
online IRL in  contexts where portions of the  observed trajectory may
be occluded.

For  differentiation, we  refer to  the  original method  as the  {\em
  batch}  version.  Recall  the  $k^{th}$ feature  expectation of  the
expert computed in Eq.~\ref{eq:latent-phi} as  part of the E-step.  If
$X_i = Y_i  \cup Z_i$ is a  trajectory in session $i$  composed of the
observed    portion   $Y_i$    and   the    hidden   portion    $Z_i$,
$\hat{\phi}^{Z|_Y,1:i}_k$  is the  expectation  computed using
all    trajectories    obtained    so     far,    we    may    rewrite
Eq.~\ref{eq:latent-phi} for feature $k$ as:

\begin{small}
\begin{align}
\hat{\phi}^{Z|_Y,1:i}_k  & \triangleq  \frac{1}{|\Ycal{}_{1:i}|}  \sum\limits_{Y
	\in   \Ycal{}_{1:i}}   \sum\limits_{Z   \in   \Z}   Pr(Z|Y;\bm{\theta})
\sum\limits_{t=1}^T \gamma^t \phi_k(\langle s, a \rangle_t)\nonumber\\
& = \frac{1}{|\Ycal{}_{1:i}|} \left ( \sum\limits_{Y
	\in   \Ycal{}_{1:i-1}}   \sum\limits_{Z   \in   \Z}   Pr(Z|Y;\bm{\theta})
\sum\limits_{t=1}^T  \gamma^t \phi_k(\langle s, a \rangle_t)+ \sum\limits_{Y
	\in   \Ycal{}_i}   \sum\limits_{Z   \in   \Z}   Pr(Z|Y;\bm{\theta}) 
\sum\limits_{t=1}^T  \gamma^t \phi_k(\langle s, a \rangle_t)\right )
   \nonumber\\
& = \frac{1}{|\Ycal{}_{1:i}|} \left (
                |\Ycal{}_{1:i-1}|~\hat{\phi}^{Z|_Y,1:i-1}_k +
                |\Ycal{}_i|~\hat{\phi}^{Z|_Y,i}_k \right )
                ~~~~~~~~~~~~~~~~~~~~~~ (\text{by definition})\nonumber\\
  & =                \frac{|\Ycal{}_{1:i-1}|}{|\Ycal{}_{1:i-1}|+|\Ycal{}_i|}~\hat{\phi}^{Z|_Y,1:i-1}_k + \frac{|\Ycal{}_i|}{|\Ycal{}_{1:i-1}|+|\Ycal{}_i|}~\hat{\phi}^{Z|_Y,i}_k
\label{eq:latent-phi-inc}
\end{align}
\end{small}

A session of our incremental LME  takes as input the expert's MDP sans
the reward function, the current session's trajectories, the number of
trajectories observed  until the start  of this session,  the expert's
empirical  feature  expectation  and   reward  weights  from  previous
session.    More    concisely,   each    session   is    denoted   by,
$\zeta_i(MDP_{/R_E},\Ycal{}_i,
|\Ycal_{1:i-1}|,\hat{\phi}^{Z|_Y,1:i-1},\bm{\theta}^{i-1})$.   In each
session,  the  feature  expectations  using  that  session's  observed
trajectories  are computed,  and the  output feature  expectations are
obtained     by    including     these    as     shown    above     in
Eq.~\ref{eq:latent-phi-inc}.   Of  course,  each session  may  involve
several iterations  of the E-  and M-steps until the  converged reward
weights $\bm{\theta}^i$  is obtained thereby giving  the corresponding
reward function estimate.  Here, the expert's feature expectation is a
sufficient statistic to compute the  reward function. We refer to this
method as LME \iirl{}.

Wang et  al.~\cite{Wang2002} shows that  if the distribution  over the
trajectories in~\eqref{eq:EM-max-ent}  is log linear, then  the reward
function  that maximizes  the entropy  of the  trajectory distribution
also  maximizes the  log likelihood  of the  observed portions  of the
trajectories.  Given  this linkage  with log likelihood,  the stopping
criterion \#1 as  given in Def.~\ref{def:stop1} is  utilized. In other
words,       the       sessions       will       terminate       when,
$|LL(\bm{\theta}^i|\Ycal{}_{1:i})                                    -
LL(\bm{\theta}^{i-1}|\Ycal{}_{1:i-1})|    \leq     \epsilon$,    where
$\bm{\theta}^i$ fully parameterizes the  reward function estimate from
the $i^{th}$ session and $\epsilon$  is a given acceptable difference.
The algorithm for this method is presented in the supplementary file.

	
 
Incremental LME admits some  significant convergence guarantees with a
confidence  of   meeting  the   desired  error-specification   on  the
demonstration likelihood. We state  these results with discussions but
defer the detailed proofs to the supplementary file.
	
\begin{lem}[Monotonicity]\label{lemma:conv_sess}Incremental
  LME increases  the demonstration likelihood monotonically  with each
  new                                                         session,
  $LL(\bm{\theta}^i|\Ycal{}_{1:i})-LL(\bm{\theta}^{i-1}|\Ycal{}_{1:i-1})
  \geqslant  0$,  when $|\Ycal{}_{1:i-1}|\gg|\Ycal{}_i|$,  yielding  a
  feasible solution to \iirl{} with missing training data.
\end{lem}

While Lemma~\ref{lemma:conv_sess} suggests that  the log likelihood of
the demonstration  can only  improve from session  to session  after a
significant amount of  training data has been  accumulated, a stronger
result illuminates  the confidence  with which  it approaches,  over a
series of sessions, the log  likelihood of the
expert's true weights $\bm{\theta}_E$.   To establish this convergence
result,  we first  focus on  the  full observability  setting.  For  a
desired bound $\varepsilon$ on  the log-likelihood loss (difference in
likelihood  w.r.t  expert's  $\bm{\theta}_E$ and  that  w.r.t  learned
$\bm{\theta}^i$) for  session $i$ in  this setting, the  confidence is
bounded as follows:
		
\begin{thm}[Confidence for incremental max-entropy \irl{}]
  \label{thm:maxent_bound}
  Given $  \Xcal{}_{1:i}$ as  the (fully observed)  demonstration till
  session $i$, $  \bm{\theta}_E \in [0,1]^K$ is  the expert's weights,
  and $\bm{\theta}^i$ is  the solution of session  $i$ for incremental
  max-entropy \irl{}, we have
  \[      LL(\bm{\theta}_E|\Xcal{}_{1:i})      -LL(\bm{\theta}^i
    |\Xcal{}_{1:i}) \leqslant \varepsilon
		\]			
		with   probability   at    least   $1-\delta$,   where
                $\delta    =    2K\exp    \left[-\frac{|\Xcal{}_{1:i}|
                    \varepsilon^2 (1-\gamma)^2}{2K^2} \right]$.
\end{thm}
	
As a  step toward relaxing  the assumption of full  observability made
above,
we first consider the error  in approximating the feature expectations
of the unobserved portions of the  data, accumulated from the first to
the current session of \iirl{}.  Notice that $\hat{\phi}^{Z|_Y,1:i}_k$
given  by  Eq.~\ref{eq:latent-phi-inc},  is an  approximation  of  the
full-observability  expectation   $\hat{\phi}^{1:i}_k$,  estimated  by
sampling            the           hidden            $Z$           from
$     Pr(Z|Y,\bm{\theta}^{i-1})$~\cite{Bogert16:Expectation}.      The
following  lemma   relates  the  error  due   to  this  sampling-based
approximation,                                                   i.e.,
$ \left|\hat{\phi}^{1:i}_k  - \hat{\phi}^{Z|_Y,1:i}_k\right|$,  to the
difference between  feature expectations  for learned policy  and that
estimated for the expert's true policy.


\begin{lem}[Constraint      Bounds       for      incremental      LME
  ]\label{lemma:cumulativeerror_bounds_latentmaxent}
  Suppose   $\Xcal{}_{1:i}$   has    portions   of   trajectories   in
  $\Z_{1:i} = \{Z|(Y,Z)\in\Xcal{}_{1:i}\}$  occluded from the learner.
  Let   $\varepsilon_{sampling}$    be   a   bound   on    the   error
  $  \left|  \hat{\phi}^{1:i}_k -  \hat{\phi}^{Z|_Y,1:i}_k  \right|_1,
  k\in  \{1,2  \ldots  K\}$  after  $N$  samples  for
  approximation.      Then,     with      probability     at     least
  $ 1-(\delta+\delta_{sampling})$, the following holds:
\[\left|E_{\X}[\phi_k] - \hat{\phi}^{Z|_Y,1:i}_k  \right|_1 \leqslant \varepsilon/2K+\varepsilon_{sampling}, k \in \{1,2 \ldots K\} \]
where      $\varepsilon,\delta$      are       as      defined      in
Theorem~\ref{thm:maxent_bound},                                    and
$\delta_{sampling}=2K\exp(-2(1-\gamma)^2    (\varepsilon_{sampling})^2
N)$.
\end{lem}
         
Theorem~\ref{thm:maxent_bound}              combined              with
Lemma~\ref{lemma:cumulativeerror_bounds_latentmaxent} now allows us to
completely relax the assumption of full observability as follows:

\begin{thm}[Confidence for incremental LME]
\label{thm:latentmaxent_bound}
Let  $\mathcal{Y}_{1:i}=\{Y|(Y,Z)\in\Xcal{}_{1:i}\}$  be the  observed
portions     of    the     demonstration     until    session     $i$,
$\varepsilon,  \varepsilon_{sampling}$   are  inputs  as   defined  in
Lemma~\ref{lemma:cumulativeerror_bounds_latentmaxent},             and
$ \bm{\theta}^i$ is  the solution of session $i$  for incremental LME,
then we have
		\[ LL(\bm{\theta}_E|\Ycal{}_{1:i}) -LL(\bm{\theta}^i|\Ycal{}_{1:i})  \leq \varepsilon_{latent}
		\]			
		with a confidence at least $ 1-\delta_{latent}$, where $\varepsilon_{latent}=\varepsilon+2K\varepsilon_{sampling}$, and
                $\delta_{latent}=\delta+\delta_{sampling}$.
\end{thm}
	
Given  required  inputs   and  fixed  $\varepsilon_{latent}$,  Theorem
~\ref{thm:latentmaxent_bound}       computes        a       confidence
$1-\delta_{latent}$.    The   amount    of   required   demonstration,
$| \mathcal{Y}_{1:i} |$, can be decided based on the desired magnitude
of the confidence.

\vspace{-0.05in}		
\section{Experiments}
\label{sec:experiment}
\vspace{-0.05in}

\begin{wrapfigure}{r}{2.0in}
\includegraphics[width=1.1in]{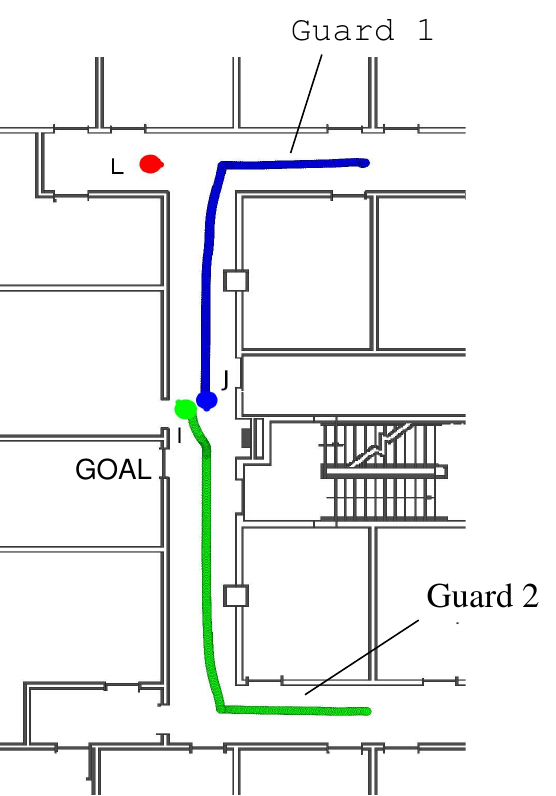} \hspace{2pt}
\includegraphics[width=0.75in]{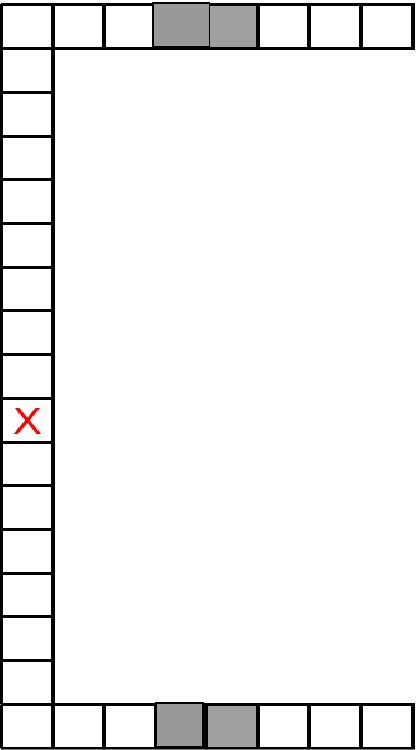}
\caption{\small The  map and  corresponding MDP  state space  for each
  patroller~\cite{Bogert14:Multi}. Shaded squares  are the turn-around
  states and the red 'X' is $L$'s goal state.  $L$ is unaware of where
  each patroller turns around or their navigation capabilities.}
\label{fig:domain}
\vspace{-0.1in}
\end{wrapfigure}

We evaluate the benefit of online  IRL on the perimeter patrol domain,
introduced by Bogert and Doshi~\cite{Bogert14:Multi}, and simulated in
ROS  using data  and files  made  publicly available.   It involves  a
robotic learner observing two guards  continuously patrol a hallway as
shown  in Fig.~\ref{fig:domain}(left).   The  learner  is tasked  with
reaching  the cell  marked 'X'  (Fig.~\ref{fig:domain}(right)) without
being spotted by  any of the patrollers.   Each guard can see  up to 3
grid cells in  front. This domain differs from  the usual applications
of IRL  toward imitation  learning.  In  particular, the  learner must
solve  its own  distinct decision-making  problem (modeled  as another
MDP) that  is reliant on knowing  how the guards patrol,  which can be
estimated  from  inferring  each   guard's  preferences.   The  guards
utilized two types of binary  state-action features leading to a total
of six:  does the current action  in the current state  make the guard
change its grid  cell?  And, is the robot's current  cell $(x,y)$ in a
given region of  the grid, which is divided into  5 regions?  The true
weight  vector   $\bm{\theta}_E$  for   these  feature   functions  is
$\langle .57, 0, 0, 0, .43, 0 \rangle$.

Notice    that    the    learner's   vantage    point    limits    its
observability. Therefore,  this domain  requires IRL  under occlusion.
Among  the  methods  discussed  in  Section~\ref{sec:background},  LME
allows IRL  when portions of  the trajectory are hidden,  as mentioned
previously.   To establish  the  benefit of  \iirl{},  we compare  the
performances of  both {\em  batch} and  {\em incremental}  variants of
this method.

\begin{figure*}[!t]
\begin{minipage}{5.5in}
  \centerline{           \includegraphics[height=1.85in,angle=-90]{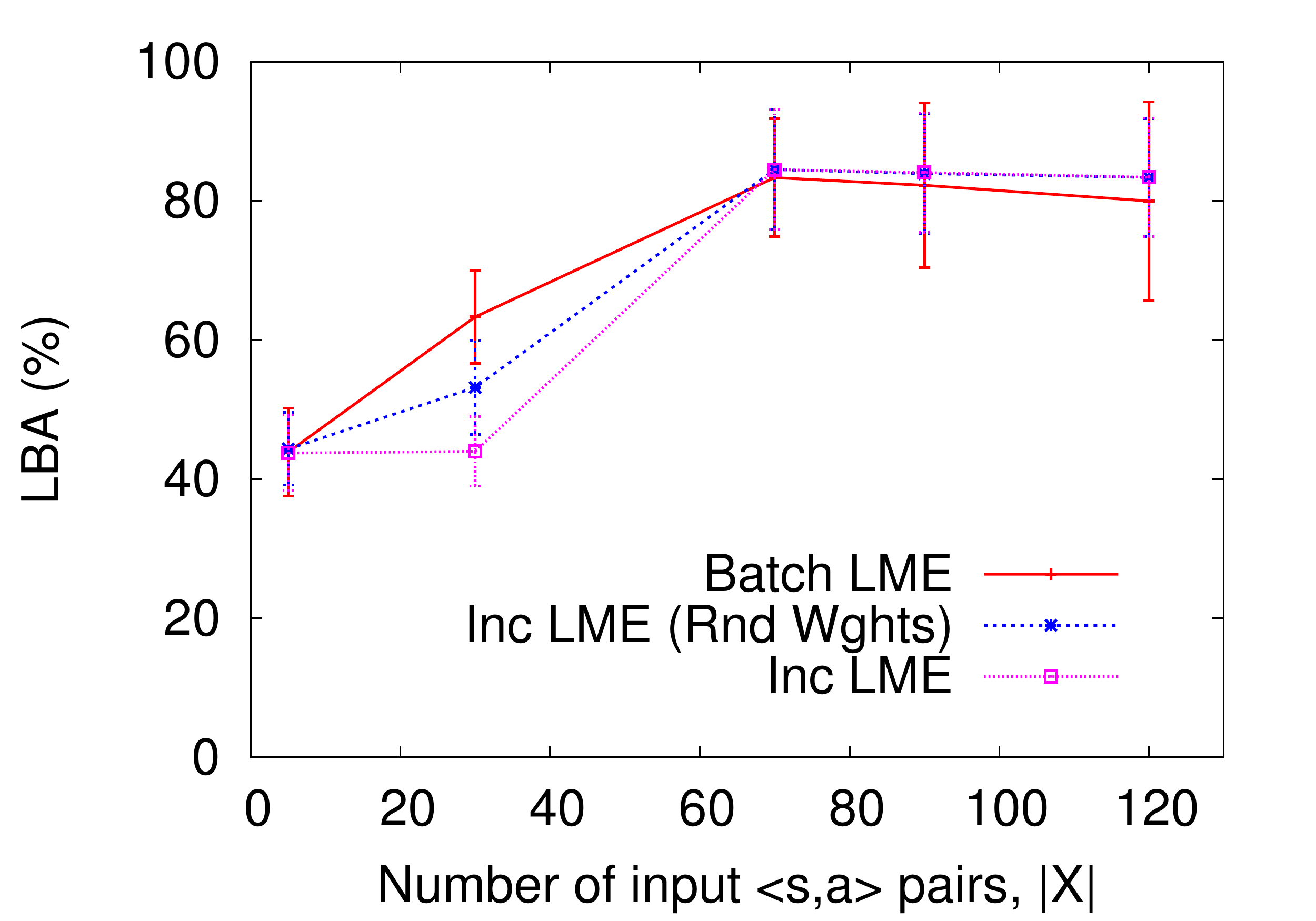}
    \includegraphics[height=1.85in,angle=-90]{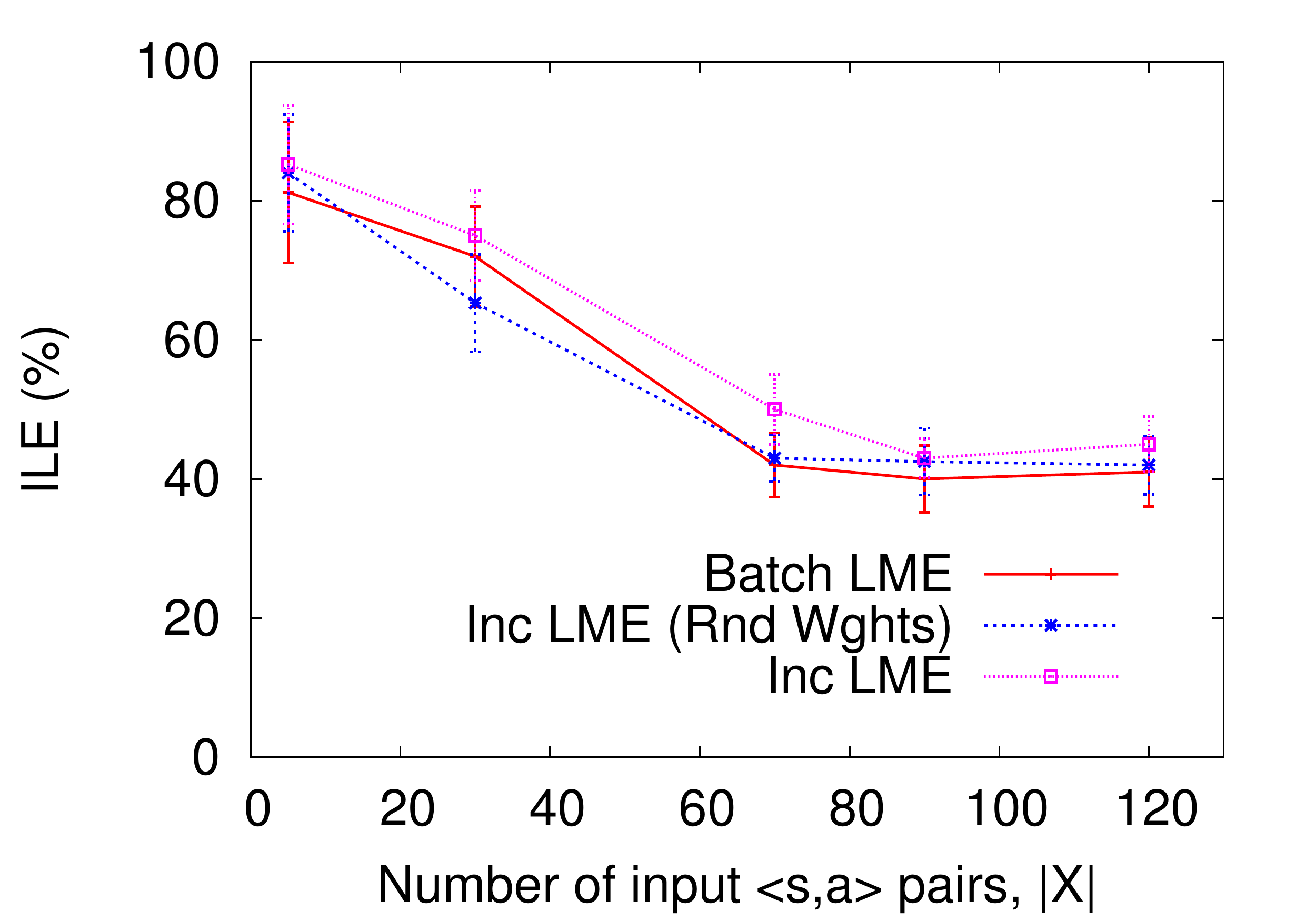}
    \includegraphics[height=1.85in,angle=-90]{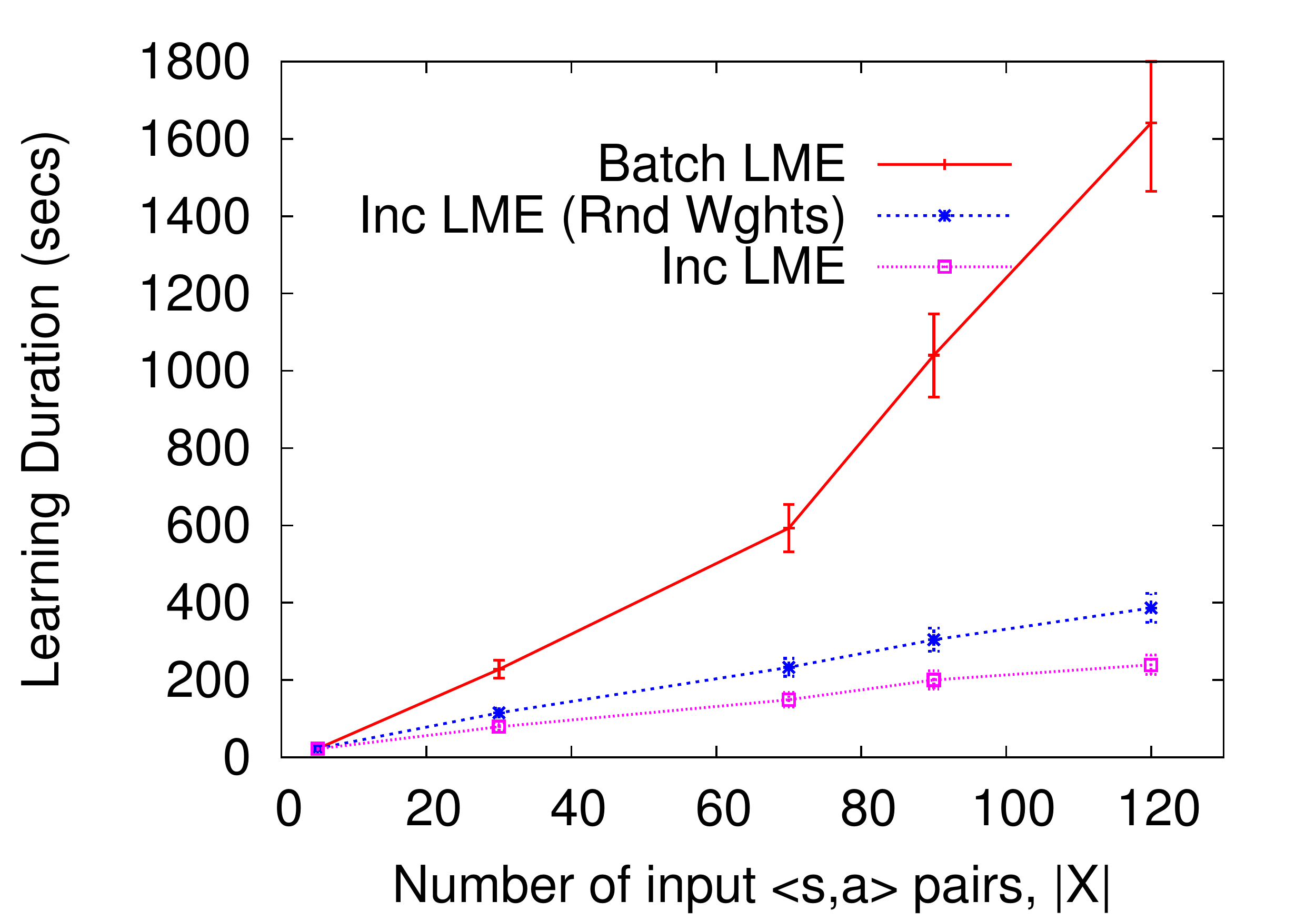}}
  \centerline{\small  \em  (a)  Learned behavior  accuracy,  ILE,  and
    learning duration under a 30\% degree of observability.}
\end{minipage}
\begin{minipage}{5.5in}
  \centerline{          \includegraphics[height=1.85in,angle=-90]{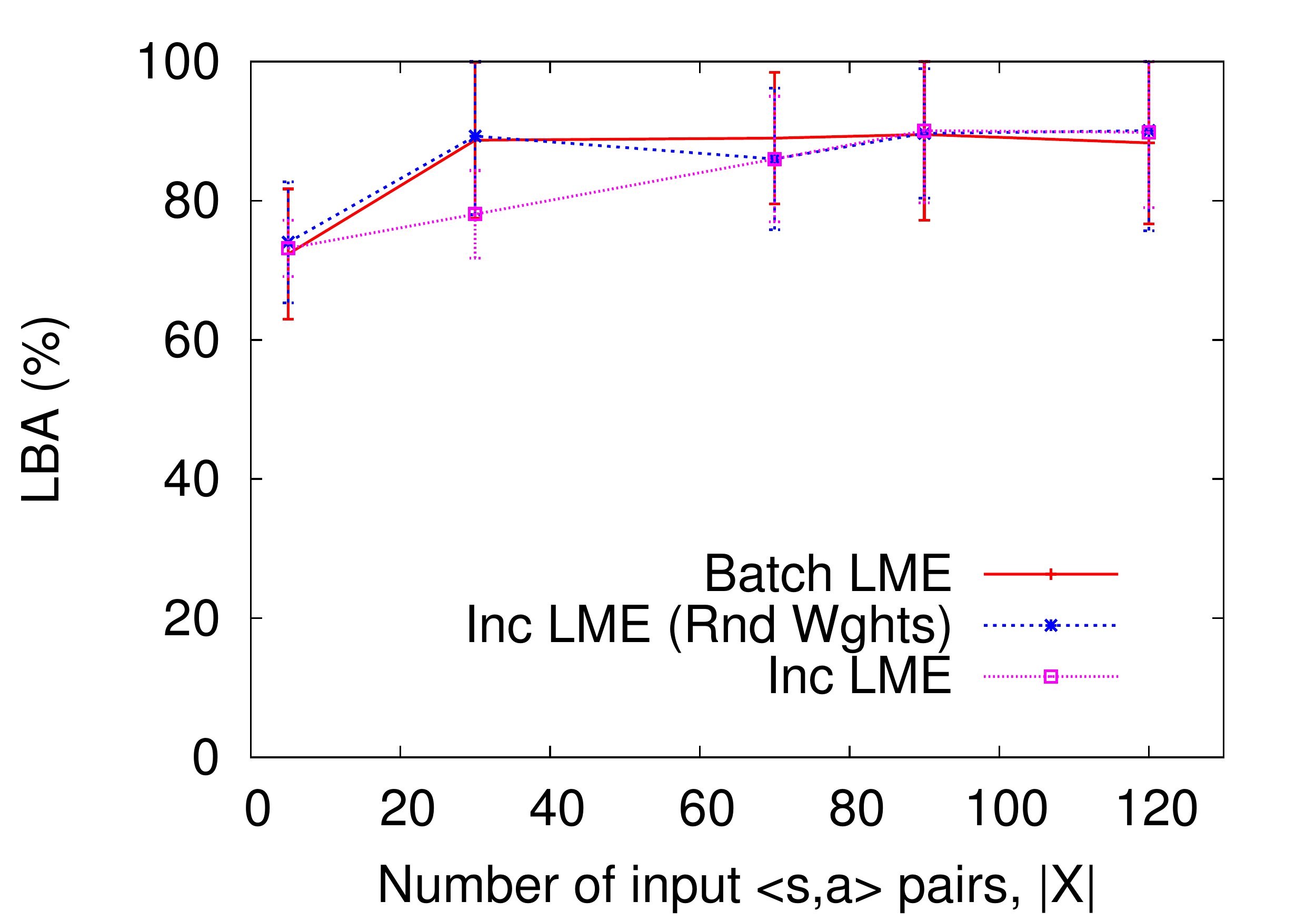}
    \includegraphics[height=1.85in,angle=-90]{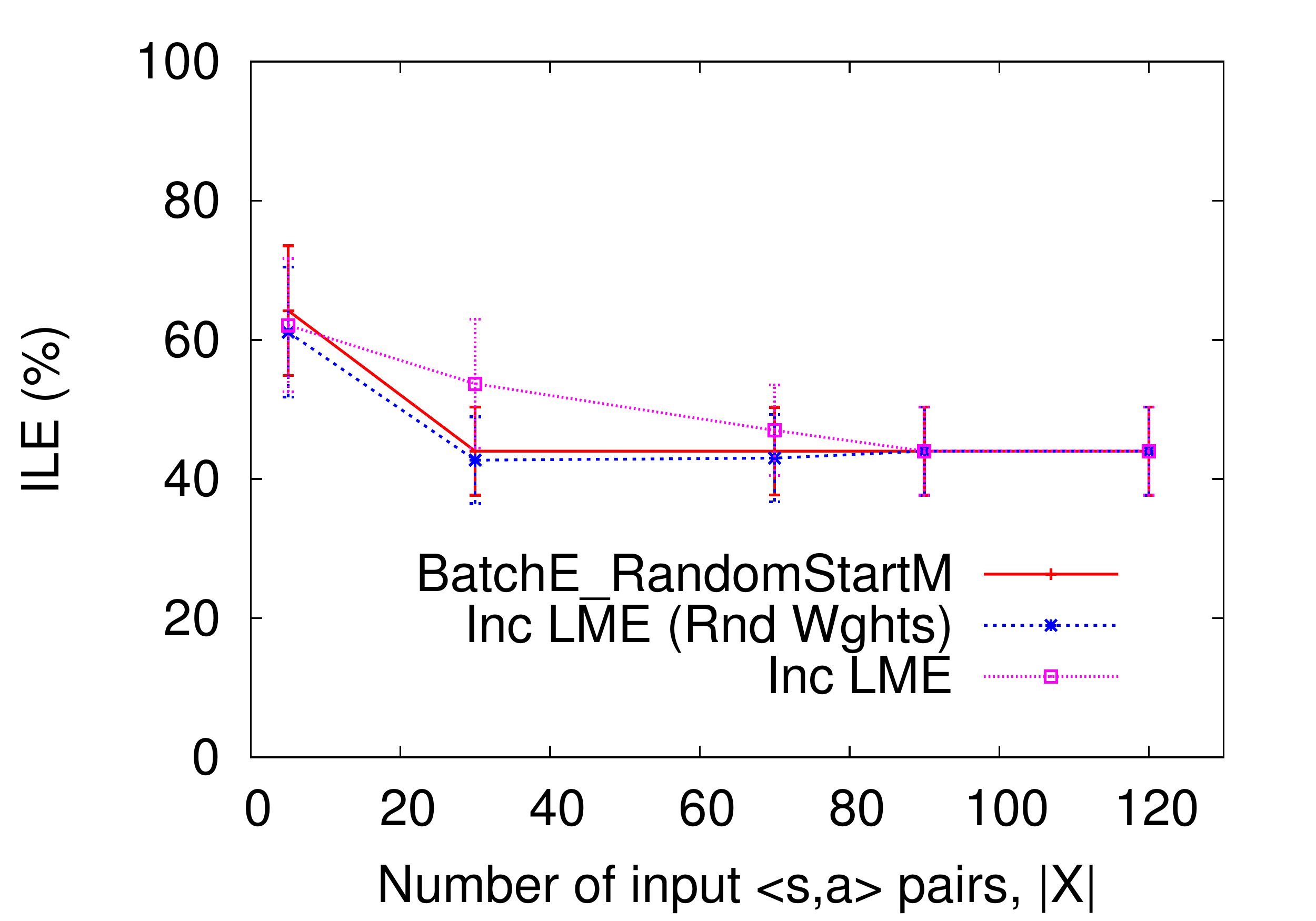}
    \includegraphics[height=1.85in,angle=-90]{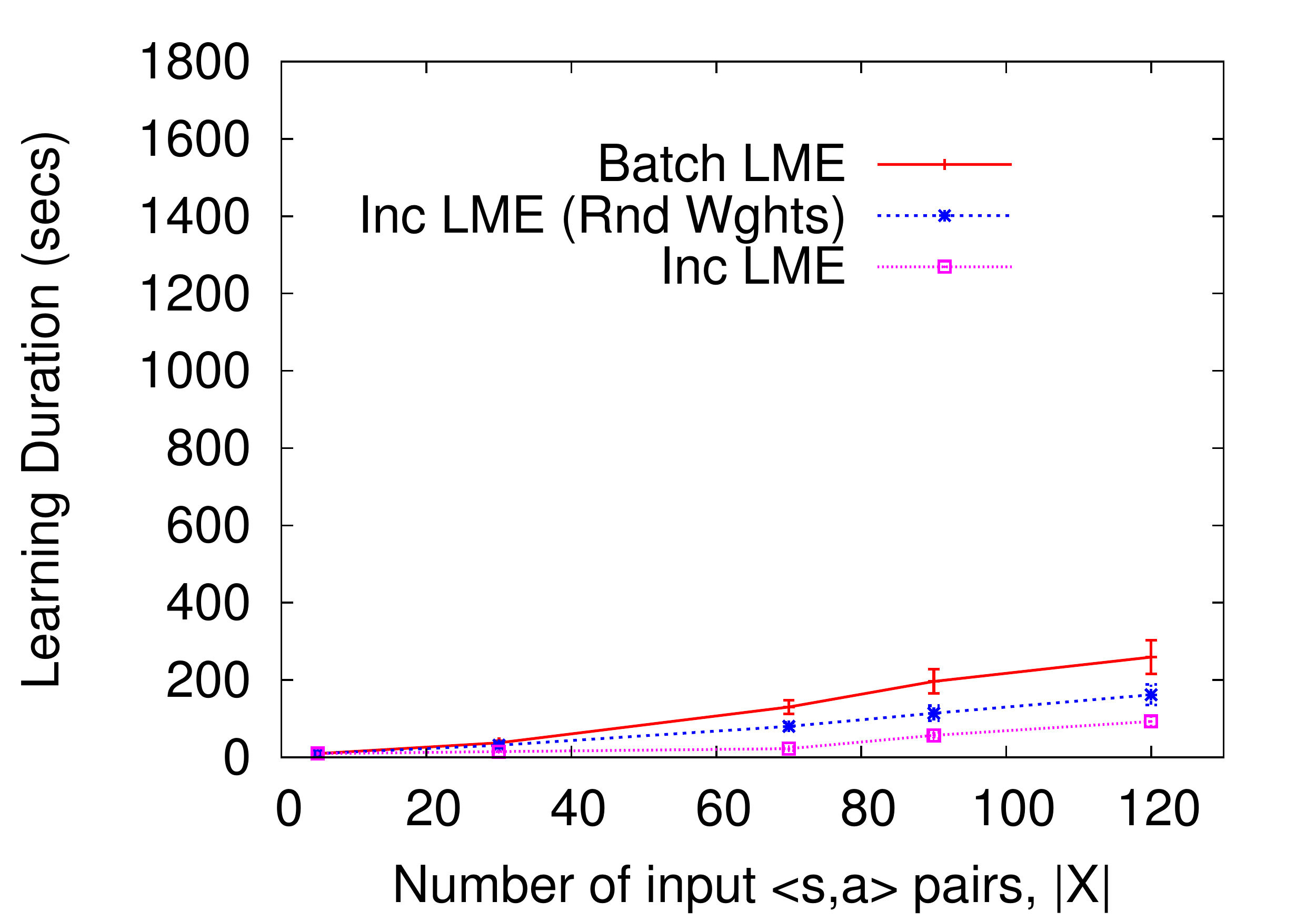}}
  \centerline{\small  \em  (b)  Learned behavior  accuracy,  ILE,  and
    learning duration under a 70\% degree of observability.}
\end{minipage}
\begin{minipage}{5.5in}
  \centerline{         \includegraphics[height=2.2in,width=1.4in,angle=-90]{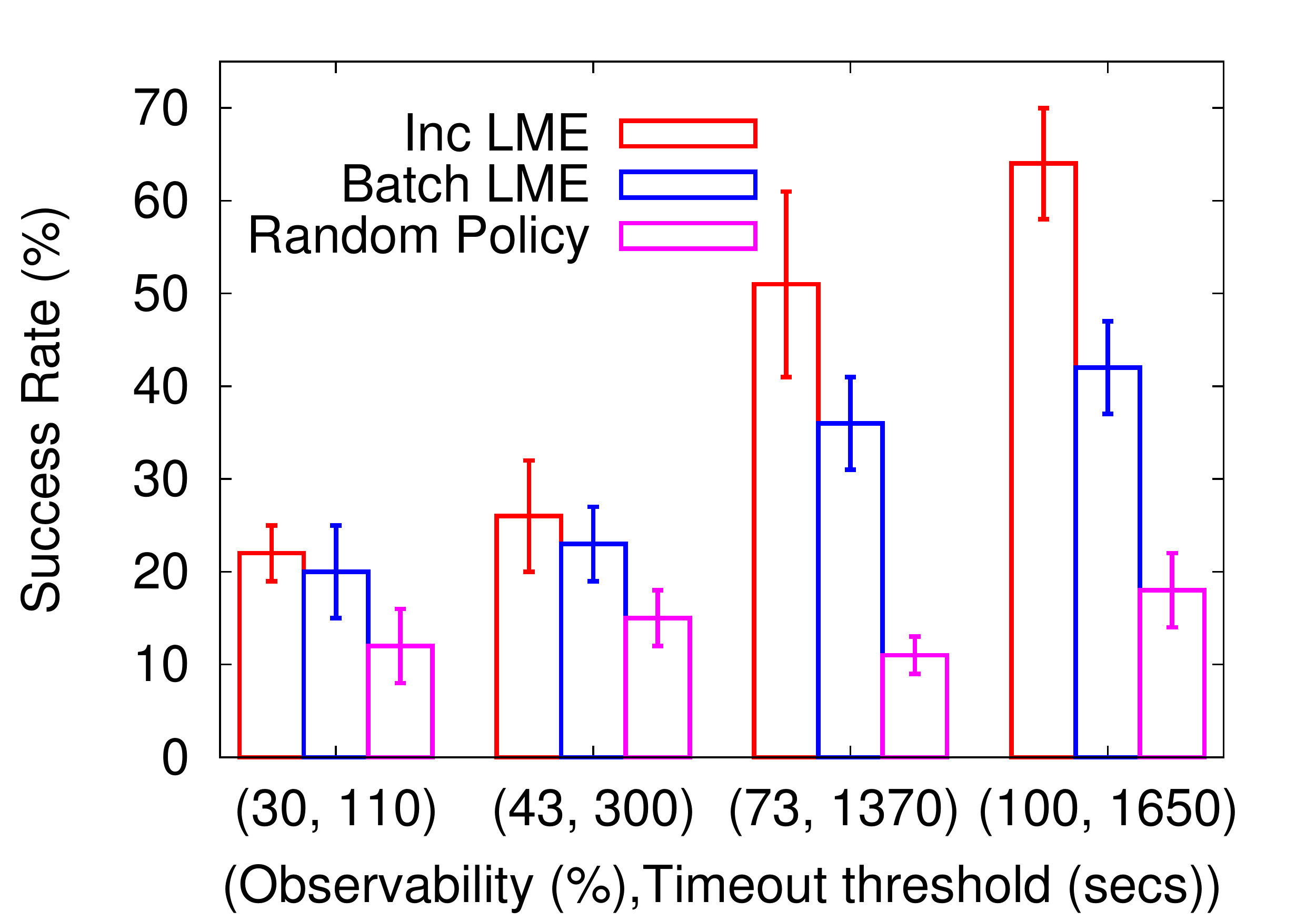}
    \includegraphics[height=2.2in,width=1.4in,angle=-90]{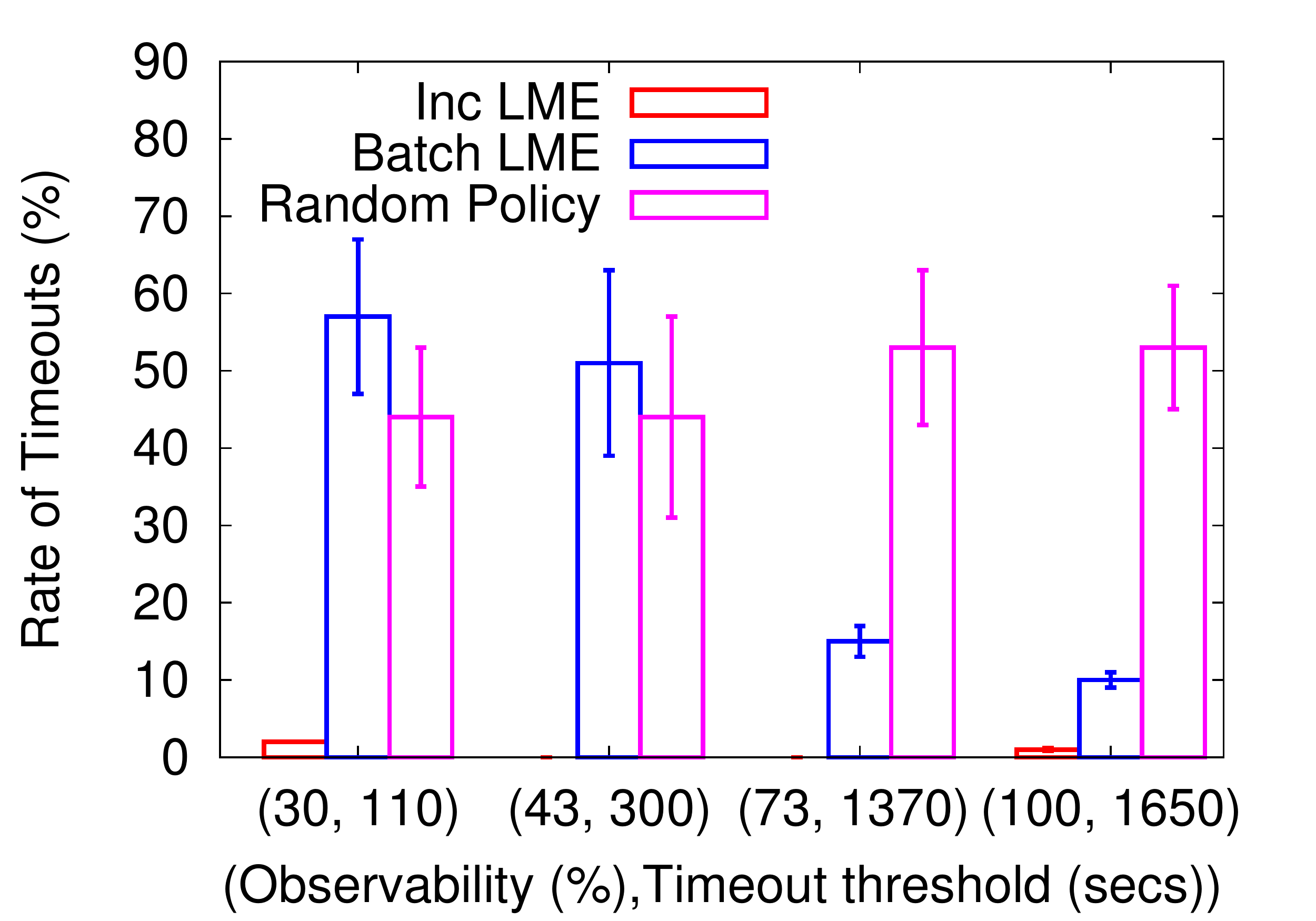}}
  {\small  \em (c)  Success rates  and timeouts  under both
    30\%, 70\%, and full observability. The success rate obtained by a
    random baseline is shown as well. This method does not perform IRL
    and picks a random time to start moving to the goal state.}
\end{minipage}
\caption{\small  Various metrics  for  comparing  the performances  of
  batch      and      incremental      LME     on      Bogert      and
  Doshi's~\cite{Bogert14:Multi}  perimeter   patrolling  domain.   Our
  experiments were  run on  a Ubuntu  16 LTS system  with an  Intel i5
  2.8GHz CPU core and 8GB RAM.}
\label{fig:learning_performances}
\vspace{-0.1in}
\end{figure*}


Efficacy of the methods was compared using the following metrics: {\em
  learned behavior accuracy}  ({\em LBA}), which is  the percentage of
overlap between the learned policy's actions and demonstrated policy's
actions;    {\em   ILE},    which    was    defined   previously    in
Section~\ref{subsec:defn};  and  {\em  success  rate},  which  is  the
percentage of runs  where $L$ reaches the goal  state undetected. Note
that when the  learned behavior accuracy is high, we  expect ILE to be
low.  However, as MDPs admit multiple optimal policies, a low ILE need
not  translate into  a high  behavior  accuracy.  As  such, these  two
metrics are not strictly correlated.
 
We  report the  LBA, ILE,  and  the time  duration in  seconds of  the
inverse   learning   for   both   batch   and   incremental   LME   in
Figs.~\ref{fig:learning_performances}$(a)$
and~\ref{fig:learning_performances}$(b)$;  the  latter  under  a  30\%
degree of observability and the former under 70\%.  Each data point is
averaged over  100 trials for  a fixed  degree of observability  and a
fixed number of state-action pairs in the demonstration $\Xcal{}$.
While the entire demonstration is given as input to the batch variant,
the  $\Xcal{}_i$ for  each session  has one  trajectory. As  such, the
incremental  learning  stops  when  there  are  no  more  trajectories
remaining to be processed.
To better understand any differentiations in performance, we introduce
a    third    variant    that     implements    each    session    as,
$\zeta_i(MDP_{/R_E},\Ycal{}_i, |\Ycal_{i:i-1}|,
\hat{\phi}^{Z|_Y,1:i-1})$. Notice that this
incremental  variant does  not utilize  the previous  session's reward
weights,  instead it  initializes them  randomly in  each session;  we
label it as {\em incremental LME with random weights}.

As the  size of  demonstration increases,  both batch  and incremental
variants exhibit  similar quality  of learning although  initially the
incremental  performs slightly  worse.   Importantly, incremental  LME
achieves these learning accuracies in significantly less time compared
to batch,  with the speed up  ratio increasing to four  as $|\Xcal{}|$
grows.


Is  there a  benefit due  to the  reduced learning  time? We  show the
success  rates of  the  learner when  each of  the  three methods  are
utilized   for   IRL   in   Fig.~\ref{fig:learning_performances}$(c)$.
Incremental  LME begins  to demonstrate  comparatively better  success
rates under 30\% observability itself, which further improves when the
observability is at 70\%.  While the batch LME's success rate does not
exceed 40\%,  the incremental  variant succeeds  in reaching  the goal
location undetected in about 65\% of the runs under full observability
(the  last  data  point).   A  deeper  analysis  to  understand  these
differences reveals that batch LME  suffers from a large percentage of
{\em timeouts}  -- more than  50\% for low observability,  which drops
down to 10\% for full observability.   A timeout occurs when IRL fails
to converge to a reward estimate in  the given amount of time for each
run.  On  the  other  hand,  incremental LME  suffers  from  very  few
timeouts. Of course, other factors play a role in success as well.


\vspace{-0.1in}
\section{Concluding Remarks}
\label{sec:conclusion}
\vspace{-0.1in}

This paper makes an important  contribution toward the nascent problem
of online IRL by offering the first formal framework, \iirl{}, to help
analyze  the  class of  methods  for  online  \irl{}. 
We  presented  a   new  method  within  the   \iirl{}  framework  that
generalizes recent advances in maximum entropy IRL to online settings.
Our comprehensive experiments show that  the new \iirl{} method indeed
improves over  the previous state-of-the-art in  time-limited domains,
by approximately  reproducing its  accuracy but in  significantly less
time. In particular, we have shown that given practical constraints on
computation time for an online IRL application, the new method suffers
fewer timeouts  and is thus  able to solve  the problem with  a higher
success rate.   In addition to  experimental validation, we  have also
established key theoretical properties of the new method, ensuring the
desired  monotonic  progress  within a  pre-computable  confidence  of
convergence.  Future  avenues for investigation  include understanding
how \iirl{} can address some  of the challenges related to scalability
to  a  larger   number  of  experts  as  well  as   the  challenge  of
accommodating unknown dynamics of the experts.



\clearpage
\bibliographystyle{abbrv}  
\bibliography{adbNIPS18}  

\begin{thebibliography}{10}

\bibitem{Abbeel04:Apprenticeship}
P.~Abbeel and A.~Y. Ng.
\newblock Apprenticeship learning via inverse reinforcement learning.
\newblock In {\em Twenty-first International Conference on Machine Learning
  (ICML)}, pages 1--8, 2004.

\bibitem{Argall09:Survey}
B.~D. Argall, S.~Chernova, M.~Veloso, and B.~Browning.
\newblock A survey of robot learning from demonstration.
\newblock {\em Robotics and autonomous systems}, 57(5):469--483, 2009.

\bibitem{Babes-Vroman11:Apprenticeship}
M.~Babes-Vroman, V.~Marivate, K.~Subramanian, and M.~Littman.
\newblock {Apprenticeship learning about multiple intentions}.
\newblock In {\em 28th International Conference on Machine Learning (ICML)},
  pages 897--904, 2011.

\bibitem{Bogert14:Multi}
K.~Bogert and P.~Doshi.
\newblock Multi-robot inverse reinforcement learning under occlusion with state
  transition estimation.
\newblock In {\em International Conference on Autonomous Agents and Multiagent
  Systems (AAMAS)}, pages 1837--1838, 2015.

\bibitem{Bogert15:Toward}
K.~Bogert and P.~Doshi.
\newblock Toward estimating others' transition models under occlusion for
  multi-robot irl.
\newblock In {\em 24th International Joint Conference on Artificial
  Intelligence (IJCAI)}, pages 1867--1873, 2015.

\bibitem{Bogert16:Expectation}
K.~Bogert, J.~F.-S. Lin, P.~Doshi, and D.~Kulic.
\newblock Expectation-maximization for inverse reinforcement learning with
  hidden data.
\newblock In {\em 2016 International Conference on Autonomous Agents and
  Multiagent Systems}, pages 1034--1042, 2016.

\bibitem{Boularias12:Structured}
A.~Boularias, O.~Kr{\"o}mer, and J.~Peters.
\newblock Structured apprenticeship learning.
\newblock In {\em European Conference on Machine Learning and Knowledge
  Discovery in Databases, Part II}, pages 227--242, 2012.

\bibitem{Choi11:Inverse}
J.~Choi and K.-E. Kim.
\newblock Inverse reinforcement learning in partially observable environments.
\newblock {\em J. Mach. Learn. Res.}, 12:691--730, 2011.

\bibitem{Dempster77:EM}
A.~P. Dempster, N.~M. Laird, and D.~B. Rubin.
\newblock Maximum likelihood from incomplete data via the em algorithm.
\newblock {\em Journal of the Royal Statistical Society, Series B
  (Methodological)}, 39:1--38, 1977.

\bibitem{Jin10:Convergence}
Z.~jun Jin, H.~Qian, S.~yi~Chen, and M.~liang Zhu.
\newblock Convergence analysis of an incremental approach to online inverse
  reinforcement learning.
\newblock {\em Journal of Zhejiang University - Science C}, 12(1):17--24, 2010.

\bibitem{Ng00:Algorithms}
A.~Ng and S.~Russell.
\newblock {Algorithms for inverse reinforcement learning}.
\newblock In {\em Seventeenth International Conference on Machine Learning},
  pages 663--670, 2000.

\bibitem{Osa18:Algorithmic}
T.~Osa, J.~Pajarinen, G.~Neumann, J.~A. Bagnell, P.~Abbeel, and J.~Peters.
\newblock An algorithmic perspective on imitation learning.
\newblock {\em Foundations and Trends® in Robotics}, 7(1-2):1--179, 2018.

\bibitem{Ramachandran07:Bayesian}
D.~Ramachandran and E.~Amir.
\newblock Bayesian inverse reinforcement learning.
\newblock In {\em 20th International Joint Conference on Artifical Intelligence
  (IJCAI)}, pages 2586--2591, 2007.

\bibitem{Rhinehart17:First}
N.~Rhinehart and K.~M. Kitani.
\newblock First-person activity forecasting with online inverse reinforcement
  learning.
\newblock In {\em International Conference on Computer Vision (ICCV)}, 2017.

\bibitem{Russell98:Learning}
S.~Russell.
\newblock Learning agents for uncertain environments (extended abstract).
\newblock In {\em Eleventh Annual Conference on Computational Learning Theory},
  pages 101--103, 1998.

\bibitem{Steinhardt14:Adaptivity}
J.~Steinhardt and P.~Liang.
\newblock Adaptivity and optimism: An improved exponentiated gradient
  algorithm.
\newblock In {\em 31st International Conference on Machine Learning}, pages
  1593--1601, 2014.

\bibitem{Wang2002}
S.~Wang, R.~Rosenfeld, Y.~Zhao, and D.~Schuurmans.
\newblock {The Latent Maximum Entropy Principle}.
\newblock In {\em IEEE International Symposium on Information Theory}, pages
  131--131, 2002.

\bibitem{Wang12:Latent}
S.~Wang and D.~{Schuurmans Yunxin Zhao}.
\newblock {The Latent Maximum Entropy Principle}.
\newblock {\em ACM Transactions on Knowledge Discovery from Data}, 6(8), 2012.

\bibitem{Ziebart08:Maximum}
B.~D. Ziebart, A.~Maas, J.~A. Bagnell, and A.~K. Dey.
\newblock Maximum entropy inverse reinforcement learning.
\newblock In {\em 23rd National Conference on Artificial Intelligence - Volume
  3}, pages 1433--1438, 2008.

\end{thebibliography}

\end{document}